\documentclass[11pt,a4paper]{article}
\usepackage[hyperref]{acl2017}
\usepackage{times}
\usepackage{latexsym}
\usepackage{url}
\usepackage{subfig}
\usepackage{graphicx}
\usepackage{forest}
\usepackage{amsmath}
\usepackage{multirow}

\aclfinalcopy 

\title{Deep Investigation of Cross-Language Plagiarism Detection Methods}

\author{
  J\'{e}r\'{e}my Ferrero \\
  Compilatio \\
  276 rue du Mont Blanc \\
  74540 Saint-F\'{e}lix, France \\
  LIG-GETALP \\
  Univ. Grenoble Alpes, France \\
  {\tt jeremy.ferrero@imag.fr} \\
  \And
  Laurent Besacier \\
  LIG-GETALP \\
  Univ. Grenoble Alpes, France \\
  {\tt laurent.besacier@imag.fr} \\
  \AND
  Didier Schwab \\
  LIG-GETALP \\
  Univ. Grenoble Alpes, France \\
  {\tt didier.schwab@imag.fr} \\
  \And  
  Fr\'{e}d\'{e}ric Agn\`{e}s \\
  Compilatio \\
  276 rue du Mont Blanc \\
  74540 Saint-F\'{e}lix, France \\
  {\tt frederic@compilatio.net} \\
}

\date{}

\begin{document}
\maketitle
\begin{abstract}
This paper is a deep investigation of cross-language plagiarism detection methods on a new recently introduced open dataset, which contains parallel and comparable collections of documents with multiple characteristics (different genres, languages and sizes of texts). 
We investigate cross-language plagiarism detection methods for 6 language pairs on 2 granularities of text units in order to draw robust conclusions on the best methods while deeply analyzing correlations across document styles and languages.
\end{abstract}

\section{Introduction}
\label{intro}

Plagiarism is a very significant problem nowadays, specifically in higher education institutions.
In monolingual context, this problem is rather well treated by several recent researches \cite{plagiarism-overview}.
Nevertheless, the expansion of the Internet, which facilitates access to documents throughout the world and to increasingly efficient (freely available) machine translation tools, helps to spread \textit{cross-language plagiarism}.
Cross-language plagiarism means plagiarism by translation, \textit{i.e.} a text has been plagiarized while being translated (manually or automatically).
The challenge in detecting this kind of plagiarism is that the suspicious document is no longer in the same language of its source.
In this relatively new field of research, no systematic evaluation of the main methods, on several language pairs, for different text granularities and for different text genres, has been proposed yet. 
This is what we propose in this paper.

\textbf{Contribution.}
The paper focus is on cross-language semantic textual similarity detection which is the main part (with source retrieval) in cross-language plagiarism detection. 
The evaluation dataset used \cite{dataset-lrec} allows us to run a large amount of experiments and analyses.
To our knowledge, this is the first time that full potential of such a diverse dataset is used for benchmarking.
So, the paper main contribution is a systematic evaluation of cross-language similarity detection methods (using in plagiarism detection) on different languages, sizes and genres of texts through a reproducible evaluation protocol.
Robust conclusions are derived on the best methods while deeply analyzing correlations across document styles and languages.
Due to space limitations, we only provide a subset of our experiments in the paper while more result tables and correlation analyses are provided as supplementary material on a Web link\footnote{\url{https://github.com/FerreroJeremy/Cross-Language-Dataset/tree/master/study}}.

\textbf{Outline.} 
After presenting the dataset used for our study in section~\ref{dataset}, and reviewing the state-of-the-art methods of cross-language plagiarism detection that we evaluate in section~\ref{methods}, we describe the evaluation protocol employed in section~\ref{protocol}.
Then, section~\ref{across} presents the correlation of the methods across language pairs, while section~\ref{detailed} presents a detailed analysis on only English-French pair. 
Finally, section~\ref{conclusion} concludes this work and gives a few perspectives.

\section{Dataset}
\label{dataset}

The reference dataset used during our study is the new dataset\footnote{\url{https://github.com/FerreroJeremy/Cross-Language-Dataset}} recently introduced by \newcite{dataset-lrec}.
The dataset was specially designed for a rigorous evaluation of cross-language textual similarity detection.
The different characteristics of the dataset are synthesized in Table~\ref{characteristics}, while Table~\ref{number} presents the number of aligned units by sub-corpus and by granularity.

\begin{table*}[t!]
\begin{center}
\begin{small}
\begin{tabular}{|l|l|l|l|l|l|}
      \hline
      \bf Sub-corpus & \bf Alignment & \bf Authors & \bf Translations & \bf Obfuscation & \bf NE (\%) \\
      \hline
      JRC-Acquis & Parallel & Politicians & Professional translators & No & 3.74 \\
      Europarl & Parallel & Politicians & Professional translators & No & 7.74 \\
      Wikipedia & Comparable & Average people & - & Noise & 8.37 \\
      PAN (Gutenberg Project) & Parallel & Professional authors & Professional authors & Yes & 3.24 \\
      Amazon Product Reviews & Parallel & Average people & Google Translate & Noise & 6.04 \\
      Conference papers & Comparable & NLP scientists & NLP scientists & Noise & 9.36 \\
      \hline
\end{tabular}
\end{small}
\end{center}
\caption{\label{characteristics} Characteristics of the dataset \cite{dataset-lrec} for each sub-corpus.
The percentages of named entities (NE) present in the last column are estimated with Stanford Named Entity Recognizer\footnotemark.}
\end{table*}

\begin{table*}[t!]
\begin{center}
\begin{small}
\begin{tabular}{|l|c|l|l|l|l|l|}
      \hline
      \bf Sub-corpus & \bf Languages & \bf \# Documents & \bf \# Sentences & \bf \# Noun chunks \\
      \hline
      JRC-Acquis & EN, FR, ES & $\simeq$ 10,000 & $\simeq$ 150,000 & $\simeq$ 10,000 \\
      Europarl & EN, FR, ES & $\simeq$ 10,000 & $\simeq$ 475,000 & $\simeq$ 25,600 \\
      Wikipedia & EN, FR, ES & $\simeq$ 10,000 & $\simeq$ 5,000 & $\simeq$ 150 \\
      PAN (Gutenberg Project) & EN, ES & $\simeq$ 3,000 & $\simeq$ 90,000 & $\simeq$ 1,400 \\
      Amazon Product Reviews & EN, FR & $\simeq$ 6,000 & $\simeq$ 23,000 & $\simeq$ 2,600 \\
      Conference papers & EN, FR & $\simeq$ 35 & $\simeq$ 1,300 & $\simeq$ 300 \\
      \hline
\end{tabular}
\end{small}
\end{center}
\caption{\label{number} Number of aligned documents, sentences and noun chunks by sub-corpus.}
\end{table*}

More precisely, the characteristics of the dataset are the following:
\begin{itemize}
\item it is multilingual: it contains French, English and Spanish texts;
\item it proposes cross-language alignment information at different granularities: document level, sentence level and chunk level;
\item it is based on both parallel and comparable corpora (mix of Wikipedia, scientific conference papers, amazon product reviews, Europarl and JRC);
\item it contains both human and machine translated texts;
\item it contains different percentages of named entities;
\item part of it has been obfuscated (to make the cross-language similarity detection more complicated) while the rest remains without noise;
\item the documents were written and translated by multiple types of authors (from average to professionals);
\item it covers various fields.
\end{itemize}

\section{Overview of State-of-the-Art Methods}
\label{methods}

\begin{figure*}[ht!]
\begin{center}
\begin{small}
\begin{forest}
    for tree={
      grow=east,
      parent anchor=south east,
      child anchor=south west,
      align=center,
      l sep+=50pt,
      edge path={
        \noexpand\path [draw, rounded corners=0pt, \forestoption{edge}] (!u.parent anchor) [out=0, in=180] to (.child anchor)\forestoption{edge label} -- (.south east);
      },
      for root={
        ellipse,
        draw,
        parent anchor=east,
      },
    }
    [
      [MT-Based Models
      	\\ \textbf{Translation + Monolingual Analysis} \cite{muhr2010}]
      [Comparable Corpora-Based Models
      	\\ \textcolor{gray}{CL-KGA}{,} \textbf{CL-ESA} \cite{potthast2008}]
      [Parallel Corpora-Based Models
      	\\ \textbf{CL-ASA} \cite{pinto2009}{,} \textcolor{gray}{CL-LSI}{,} \textcolor{gray}{CL-KCCA}]
      [Dictionary-Based Models
      	\\ \textcolor{gray}{CL-VSM}{,} \textbf{CL-CTS} \cite{pataki2012}]
      [Syntax-Based Models
      	\\ \textcolor{gray}{Length Model} \cite{pouliquen2003a}{,} \textbf{CL-C$n$G} \cite{potthast2011}{,} \textcolor{gray}{Cognateness}]
      ]
    ]
\end{forest}
\end{small}
\end{center}
\caption{\label{tree} Taxonomy of \newcite{potthast2011}, enriched by the study of \newcite{danilova2013}, of different approaches for cross-language similarity detection.}
\end{figure*}
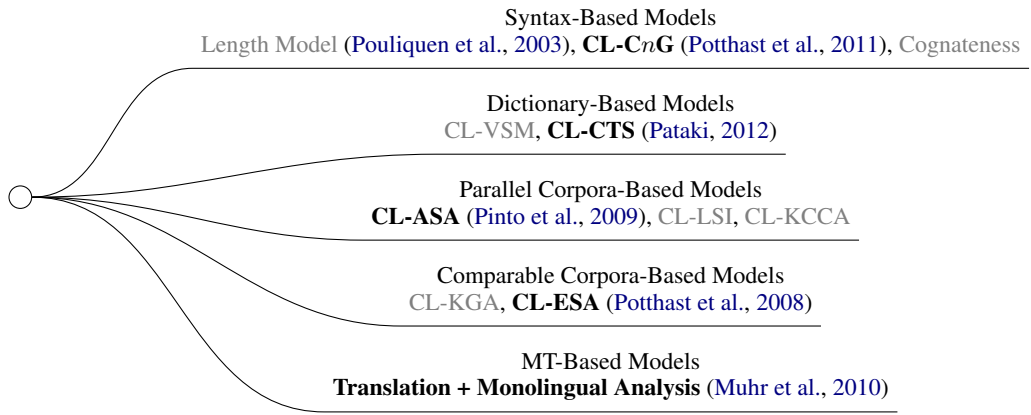

Textual similarity detection methods are not exactly methods to detect plagiarism.
Plagiarism is a statement that someone copied text deliberately without attribution, while these methods only detect textual similarities.
There is no way of knowing why texts are similar and thus to assimilate these similarities to plagiarism.
\\
\\
At the moment, there are five classes of approaches for cross-language plagiarism detection.
The aim of each method is to estimate if two textual units in different languages express the same message or not.
Figure~\ref{tree} presents a taxonomy of \newcite{potthast2011}, enriched by the study of \newcite{danilova2013}, of the different cross-language plagiarism detection methods grouped by class of approaches.
We only describe below the state-of-the-art methods that we evaluate in the paper, one for each class of approaches (those in bold in the Figure~\ref{tree}).

\paragraph{Cross-Language Character \mbox{N-Gram} (\mbox{CL-C$n$G})} is based on \newcite{mcnamee2004} model. 
We use the \emph{\mbox{CL-C3G}} \newcite{potthast2011}'s implementation.
Only spaces and alphanumeric characters are kept.
Any other diacritic or symbol is deleted and the texts are lower-cased.
The texts are then segmented into \mbox{$3$-grams} (sequences of 3 contiguous characters) and transformed into \textit{tf.idf} vectors of character \mbox{$3$-grams}.
The metric used to compare two vectors is the cosine similarity.

\paragraph{Cross-Language Conceptual Thesaurus-based Similarity (\mbox{CL-CTS})} aims to measure the semantic similarity using abstract concepts from words in textual units.
We reuse the idea of \newcite{pataki2012} which, for each sentence, build a bag-of-words by getting all the available translations of each word of the sentence.
For that, we use a linked lexical resource called \mbox{\textit{DBNary}} \cite{dbnary}.
The bag-of-words of a sentence is the merge of the bag-of-words of the words of the sentence.
After, we use the Jaccard distance \citep{jaccard1912} with fuzzy matching between two bag-of-words to measure the similarity between two sentences.

\paragraph{Cross-Language Alignment-based Similarity Analysis (\mbox{CL-ASA})} was introduced for the first time by \newcite{barron2008} and developed subsequently by \newcite{pinto2009}.
The model aims to determinate how a textual unit is potentially the translation of another textual unit using bilingual unigram dictionary which contains translations pairs (and their probabilities) extracted from a parallel corpus.
Our lexical dictionary is calculated applying the \mbox{IBM-1} model \cite{brown1993} on the concatenation of 
\footnotetext{\url{http://nlp.stanford.edu/software/CRF-NER.shtml}} 
TED\footnote{\url{https://wit3.fbk.eu/}} \cite{EAMT2012} and News\footnote{\url{http://www.statmt.org/wmt13/translation-task.html\#download}} parallel corpora.
We reuse the implementation of \newcite{pinto2009} that proposed a formula that factored the alignment function.

\paragraph{Cross-Language Explicit Semantic Analysis (\mbox{CL-ESA})} is based on the explicit semantic analysis model introduced for the first time by \newcite{gabrilovich2007}, which represents the meaning of a document by a vector based on the vocabulary derived from \mbox{Wikipedia}, to find a document within a corpus.
It was reused by \newcite{potthast2008} in the context of cross-language document retrieval.
Our implementation uses a part of Wikipedia, from which our test data was removed, to build the vector representations of the texts.

\paragraph{Translation + Monolingual Analysis (\mbox{T+MA})} consists in translating suspect plagiarized text back into the same language of source text, in order to operate a monolingual comparison between them.
We use the \newcite{muhr2010}'s implementation which consists in replacing each word of one text by its most likely translations in the language of the other text, leading to a bags-of-words.
We use \mbox{\textit{DBNary}} \cite{dbnary} to get the translations.
The metric used to compare two texts is a monolingual matching based on strict intersection of bags-of-words.
\\
\\
More recently, \mbox{SemEval-2016} \cite{semeval2016} proposed a new subtask on evaluation of cross-lingual semantic textual similarity.
Despite the fact that it was the first year that this subtask was attempted, there were 26 submissions from 10 teams.
Most of the submissions relied on a machine translation step followed by a monolingual semantic similarity, but 4 teams tried to use learned vector representations (on words or sentences) combined with machine translation confidence (for instance the submission of \newcite{cnrc} or \newcite{fbk}).
The method that achieved the best performance \cite{uwb} was a supervised system built on a word alignment-based method proposed by \newcite{sultan}.
This very recent method is, however, not evaluated in this paper.

\begin{table*}[ht!]
 \begin{center}
\begin{small}
\begin{tabular}{|l|l|l|l|l|l|l|}
      \hline
	  \multicolumn{7}{|c|}{\bf Chunk level} \\
      \hline
      \bf Methods & \bf EN$\rightarrow$FR & \bf FR$\rightarrow$EN & \bf EN$\rightarrow$ES & \bf ES$\rightarrow$EN & \bf ES$\rightarrow$FR & \bf FR$\rightarrow$ES \\
      \hline
      CL-C3G & \textbf{0.5071} & \textbf{0.5071} & \textbf{0.4375} & \textbf{0.4375} & \textbf{0.4795} & \textbf{0.4795} \\
      CL-CTS & 0.4250 & 04116 & 0.3780 & 0.3881 & 0.4203 & 0.4169 \\
      CL-ASA & 0.4738 & 0.4252 & 0.4083 & 0.3941 & 0.3736 & 0.3540 \\
      CL-ESA & 0.1499 & 0.1499 & 0.1476 & 0.1476 & 0.1520 & 0.1520 \\
      T+MA & 0.3730 & 0.3634 & 0.3177 & 0.3279 & 0.3158 & 0.3140 \\
	  \hline
	  \hline
	  \multicolumn{7}{|c|}{\bf Sentence level} \\
      \hline
      \bf Methods & \bf EN$\rightarrow$FR & \bf FR$\rightarrow$EN & \bf EN$\rightarrow$ES & \bf ES$\rightarrow$EN & \bf ES$\rightarrow$FR & \bf FR$\rightarrow$ES \\
      \hline
      CL-C3G & \textbf{0.4931} & \textbf{0.4931} & \textbf{0.3819} & \textbf{0.3819} & 0.4577 & \textbf{0.4577} \\
      CL-CTS & 0.4734 & 0.4633 & 0.3171 & 0.3204 & \textbf{0.4645} & 0.4575 \\
      CL-ASA & 0.3576 & 0.3523 & 0.2694 & 0.2531 & 0.3098 & 0.2843 \\
      CL-ESA & 0.1430 & 0.1430 & 0.1337 & 0.1337 & 0.1383 & 0.1383 \\
      T+MA & 0.3760 & 0.3692 & 0.3505 & 0.3526 & 0.3673 & 0.3525 \\
      \hline
\end{tabular}
\end{small}
 \end{center}
\caption{\label{correl_lang} Overall~$F_{1}$~score over all sub-corpora of the state-of-the-art methods for each language pair (EN: English; FR: French; ES: Spanish).}
\end{table*}

\begin{table*}[ht!]
 \begin{center}
\begin{small}
\begin{tabular}{|l|l|l|l|l|l||l|l|}
      \hline
	  \multicolumn{8}{|c|}{\bf Chunk level} \\
      \hline
      \bf EN$\rightarrow$FR & \bf FR$\rightarrow$EN & \bf EN$\rightarrow$ES & \bf ES$\rightarrow$EN & \bf ES$\rightarrow$FR & \bf FR$\rightarrow$ES & \bf Overall & \bf Lang. Pair  \\
      \hline
      1.000 & 0.991 & \textbf{0.998} & 0.995 & 0.957 & 0.940 & 0.980 & \textbf{\text{EN$\rightarrow$FR}} \\
      \cline{1-8}
      \multicolumn{1}{r|}{} & 1.000 & 0.990 & 0.994 & 0.980 & 0.971 & 0.987 & \textbf{\text{FR$\rightarrow$EN}} \\
      \cline{2-8}
      \multicolumn{2}{r|}{} & 1.000 & 0.996 & 0.967 & 0.949 & 0.983 & \textbf{\text{EN$\rightarrow$ES}} \\
      \cline{3-8}
      \multicolumn{3}{r|}{} & 1.000 & 0.978 & 0.965 & 0.988 & \textbf{\text{ES$\rightarrow$EN}} \\
      \cline{4-8}
      \multicolumn{4}{r|}{} & 1.000 & \textbf{0.998} & 0.980 & \textbf{\text{ES$\rightarrow$FR}} \\
      \cline{5-8}
      \multicolumn{5}{r|}{} & 1.000 & 0.970 & \textbf{\text{FR$\rightarrow$ES}} \\
      \cline{6-8}
      \multicolumn{8}{r}{} \\
      \hline
	  \multicolumn{8}{|c|}{\bf Sentence level} \\
      \hline
      \bf EN$\rightarrow$FR & \bf FR$\rightarrow$EN & \bf EN$\rightarrow$ES & \bf ES$\rightarrow$EN & \bf ES$\rightarrow$FR & \bf FR$\rightarrow$ES & \bf Overall & \bf Lang. Pair  \\
      \hline
      1.000 & 1.000 & 0.929 & 0.922 & 0.991 & 0.982 & 0.971 & \textbf{\text{EN$\rightarrow$FR}} \\
      \cline{1-8}
      \multicolumn{1}{r|}{} & 1.000 & 0.931 & 0.924 & 0.989 & 0.981 & 0.971 & \textbf{\text{FR$\rightarrow$EN}} \\
      \cline{2-8}
      \multicolumn{2}{r|}{} & 1.000 & \textbf{0.997} & 0.925 & 0.913 & 0.949 & \textbf{\text{EN$\rightarrow$ES}} \\
      \cline{3-8}
      \multicolumn{3}{r|}{} & 1.000 & 0.928 & 0.922 & 0.949 & \textbf{\text{ES$\rightarrow$EN}} \\
      \cline{4-8}
      \multicolumn{4}{r|}{} & 1.000 & \textbf{0.997} & 0.971 & \textbf{\text{ES$\rightarrow$FR}} \\
      \cline{5-8}
      \multicolumn{5}{r|}{} & 1.000 & 0.966 & \textbf{\text{FR$\rightarrow$ES}} \\
      \cline{6-8}
\end{tabular}
\end{small}
 \end{center}
\caption{\label{correl_by_lang} Pearson correlations of the overall~$F_{1}$~score over all sub-corpora of all methods between the different language pairs (EN: English; FR: French; ES: Spanish).}
\end{table*}

\section{Evaluation Protocol}
\label{protocol}

We apply the same evaluation protocol as in \newcite{dataset-lrec}'s paper.
We build a distance matrix of size~$N$\,x\,$M$, with~$M$~=~1,000 and~$N$~=~$|S|$ where~$S$ is the evaluated sub-corpus.
Each textual unit of~$S$ is compared to itself (actually, since this is cross-lingual similarity detection, each source language unit is compared to its corresponding unit in the target language) and to~$M$-1 other units randomly selected from~$S$.
The same unit may be selected several times.
Then, a matching score for each comparison performed is obtained, leading to the distance matrix.
Thresholding on the matrix is applied to find the threshold giving the best~$F_{1}$~score.
The~$F_{1}$~score is the harmonic mean of precision and recall.
Precision is defined as the proportion of relevant matches (similar cross-language units) retrieved among all the matches retrieved.
Recall is the proportion of relevant matches retrieved among all the relevant matches to retrieve.
Each method is applied on each sub-corpus for chunk and sentence granularities.
For each configuration (\textit{i.e.} a particular method applied on a particular sub-corpus considering a particular granularity), 10 folds are carried out by changing the~$M$ selected units.

\section{Investigation of Cross-Language Similarity Performances}
\label{expe}

\subsection{Across Language Pairs}
\label{across}

Table~\ref{correl_lang} brings together the performances of all methods on all sub-corpora for each pair of languages at \textit{chunk} and \textit{sentence} level.
In both sub-tables, at chunk and sentence level, the overall~$F_{1}$~score over all sub-corpora of one method in one particular language pair is given. 

As a preliminary remark, one should note that \emph{\mbox{CL-C3G}} and \emph{\mbox{CL-ESA}} lead to the same results for a given language pair (same performance if we reverse source and target languages) due to their symmetrical property.
Another remark we can make is that methods are consistent across language pairs: best performing methods are mostly the same, whatever the language pair considered. 
This is confirmed by the calculation of the Pearson correlation between performances of different pairs of languages, from Table~\ref{correl_lang} and reported in Table~\ref{correl_by_lang}.
Table~\ref{correl_by_lang} represents the Pearson correlations between the different language pairs of the overall results of all methods on all sub-corpora.
This result is interesting because some of these methods depend on the availability of lexical resources whose quality is heterogeneous across languages.
Despite the variation of the source and target languages, a minimum Pearson correlation of 0.940 for EN$\rightarrow$FR vs. FR$\rightarrow$ES, and a maximum of 0.998 for EN$\rightarrow$FR vs. EN$\rightarrow$ES and ES$\rightarrow$FR vs. FR$\rightarrow$ES at chunk level is observed (see Table~\ref{correl_by_lang}).
For the sentence granularity, it is the same order of magnitude: the maximum Pearson correlation is 0.997 for ES$\rightarrow$EN vs. EN$\rightarrow$ES and ES$\rightarrow$FR vs. FR$\rightarrow$ES, and the minimum is 0.913 for EN$\rightarrow$ES vs. FR$\rightarrow$ES (see Table~\ref{correl_by_lang}).
In average the language pair EN$\rightarrow$FR is 0.975 correlated with the other language pairs (0.980 at chunk-level and 0.971 at sentence-level), for instance.
This correlation suggests the possibility to tune a method on one language and apply it to another language if needed.

Table~\ref{top_methods} synthesizes the top~3 methods for each language pair observed in Tables~\ref{correl_lang} and~\ref{correl_by_lang}.
No matter the source and target languages or the granularity, \emph{\mbox{CL-C3G}} generally outperforms the other methods.
Then \emph{\mbox{CL-ASA}}, \emph{\mbox{CL-CTS}} and \emph{\mbox{T+MA}} are also closely efficient but their behavior depends on the granularity.
Generally, \emph{\mbox{CL-ASA}} is better at the chunk granularity, followed by \emph{\mbox{CL-CTS}} and \emph{\mbox{T+MA}}.
On the contrary, \emph{\mbox{CL-CTS}} and \emph{\mbox{T+MA}} are slightly more effective at sentence granularity.
One explanation for this is that \emph{\mbox{T+MA}} depends on the quality of machine translation, which may have poor performance on isolated chunks, while a short length text unit benefits the \emph{\mbox{CL-CTS}} and \emph{\mbox{CL-ASA}} methods because of their formula which will tend to minimize the number of false positives in this case. 
Anyway, despite these differences in ranking, the gap in term of performance values is small between these closest methods.
For instance, we can see that when \emph{\mbox{CL-CTS}} is more efficient than \emph{\mbox{CL-C3G}} (ES$\rightarrow$FR column at sentence level in Table~\ref{correl_lang} and Table~\ref{top_methods} (b)), the difference of performance is very small (0.0068).

\begin{table}[t!]
  \centering
  \subfloat[Chunk granularity]{
\begin{small}
\begin{tabular}{|l|l|}
      \hline
      \bf EN$\leftrightarrow$FR & \bf  ES$\leftrightarrow$FR \\
      \bf EN$\leftrightarrow$ES & \\
      \hline
      CL-C3G & CL-C3G \\
      CL-ASA & CL-CTS \\
      CL-CTS & CL-ASA \\
      \hline
\end{tabular}
\end{small}
  }\hspace{1cm}
  \subfloat[Sentence granularity]{
\begin{small}
\begin{tabular}{|l|l|l|}
      \hline
      \bf EN$\leftrightarrow$FR & \bf EN$\leftrightarrow$ES & \bf  ES$\rightarrow$FR \\
      \bf FR$\rightarrow$ES & & \\
      \hline
      CL-C3G & CL-C3G & CL-CTS \\
      CL-CTS & T+MA & CL-C3G \\
      T+MA & CL-CTS & T+MA \\
      \hline
\end{tabular}
\end{small}
  }
  \caption{\label{top_methods} Top 3 methods by source and target language.}
\end{table}

\begin{table}[t!]
 \begin{center}
\begin{small}
\begin{tabular}{|l|l|}
      \hline
      \bf Lang. Pair & \bf Correlation \\
      \hline
      EN$\rightarrow$FR & 0.907 \\
      FR$\rightarrow$EN & 0.946 \\
      EN$\rightarrow$ES & 0.833 \\
      ES$\rightarrow$EN & 0.838 \\
      ES$\rightarrow$FR & 0.932 \\
      FR$\rightarrow$ES & 0.939 \\
      \hline
\end{tabular}
\end{small}
 \end{center}
\caption{\label{meth_gran} Pearson correlations of the results of all methods on all sub-corpora, between the chunk and the sentence granularity, by language pair (EN: English; FR: French; ES: Spanish) (calculated from Table~\ref{correl_lang}).}
\end{table}

\begin{table}[t!]
 \begin{center}
\begin{small}
\begin{tabular}{|l|l|}
      \hline
      \bf Methods & \bf Correlation \\
      \hline
      CL-C3G & 0.996 \\
      CL-CTS &  0.970 \\
      CL-ASA & 0.649 \\
      CL-ESA & 0.515 \\
      T+MA & 0.780 \\
      \hline
\end{tabular}
\end{small}
 \end{center}
\caption{\label{lang_gran} Pearson correlations of the results on all sub-corpora on all language pairs, between the chunk and the sentence granularity, by methods (calculated from Table~\ref{correl_lang}).}
\vspace{-1em}
\end{table}

\begin{table*}[t!]
 \begin{center}
\begin{small}
\begin{tabular}{|l|l|l|l|l|l||l|}
	  \hline
	  \multicolumn{7}{|c|}{\bf Chunk level} \\
      \hline
      \bf Methods & \bf Wikipedia (\%) & \bf TALN (\%) & \bf JRC (\%) & \bf APR (\%) & \bf Europarl (\%) & \bf Overall (\%) \\
      \hline
      CL-C3G & 62.91 {\scriptsize $\pm$ 0.815} & 40.90 {\scriptsize $\pm$ 0.500} & 36.63 {\scriptsize $\pm$ 0.826} & 80.30 {\scriptsize $\pm$ 0.703} & 53.29 {\scriptsize $\pm$ 0.583} & 50.71 {\scriptsize $\pm$ 0.655} \\
      CL-CTS & 58.00 {\scriptsize $\pm$ 0.519} & 33.71 {\scriptsize $\pm$ 0.382} & 29.87 {\scriptsize $\pm$ 0.815} & 67.51 {\scriptsize $\pm$ 1.050} & 44.95 {\scriptsize $\pm$ 1.157} & 42.50 {\scriptsize $\pm$ 1.053} \\
      CL-ASA & 23.33 {\scriptsize $\pm$ 0.724} & 23.39 {\scriptsize $\pm$ 0.432} & 33.14 {\scriptsize $\pm$ 0.936} & 26.49 {\scriptsize $\pm$ 1.205} & 55.50 {\scriptsize $\pm$ 0.681} & 47.38 {\scriptsize $\pm$ 0.781} \\
      CL-ESA & 64.89 {\scriptsize $\pm$ 0.664} & 23.78 {\scriptsize $\pm$ 0.613} & 14.03 {\scriptsize $\pm$ 0.997} & 23.14 {\scriptsize $\pm$ 0.777} & 14.19 {\scriptsize $\pm$ 0.590} & 14.99 {\scriptsize $\pm$ 0.709} \\
      T+MA & 58.22 {\scriptsize $\pm$ 0.756} & 39.13 {\scriptsize $\pm$ 0.551} & 28.61 {\scriptsize $\pm$ 0.597} & 73.14 {\scriptsize $\pm$ 0.666} & 36.95 {\scriptsize $\pm$ 1.502} & 37.30 {\scriptsize $\pm$ 1.200} \\
      \hline
      \hline
      \multicolumn{7}{|c|}{\bf Sentence level} \\
      \hline
      \bf Methods & \bf Wikipedia (\%) & \bf TALN (\%) & \bf JRC (\%) & \bf APR (\%) & \bf Europarl (\%) & \bf Overall (\%) \\
      \hline
      CL-C3G & 48.25 {\scriptsize $\pm$ 0.349} & 48.08 {\scriptsize $\pm$ 0.538} & 36.68 {\scriptsize $\pm$ 0.693} & 61.10 {\scriptsize $\pm$ 0.581} & 52.72 {\scriptsize $\pm$ 0.866} & 49.31 {\scriptsize $\pm$ 0.798} \\
      CL-CTS & 46.68 {\scriptsize $\pm$ 0.437} & 38.67 {\scriptsize $\pm$ 0.552} & 28.21 {\scriptsize $\pm$ 0.612} & 50.82 {\scriptsize $\pm$ 1.034} & 53.21 {\scriptsize $\pm$ 0.601} & 47.34 {\scriptsize $\pm$ 0.632} \\
      CL-ASA & 27.63 {\scriptsize $\pm$ 0.330} & 27.25 {\scriptsize $\pm$ 0.341} & 35.17 {\scriptsize $\pm$ 0.644} & 25.53 {\scriptsize $\pm$ 0.795} & 36.55 {\scriptsize $\pm$ 1.139} & 35.76 {\scriptsize $\pm$ 0.978} \\
      CL-ESA & 51.14 {\scriptsize $\pm$ 0.875} & 14.25 {\scriptsize $\pm$ 0.334} & 14.44 {\scriptsize $\pm$ 0.341} & 13.93 {\scriptsize $\pm$ 0.714} & 13.91 {\scriptsize $\pm$ 0.618} & 14.30 {\scriptsize $\pm$ 0.551} \\
      T+MA & 50.57 {\scriptsize $\pm$ 0.888} & 37.79 {\scriptsize $\pm$ 0.364} & 32.36 {\scriptsize $\pm$ 0.369} & 61.94 {\scriptsize $\pm$ 0.756} & 37.92 {\scriptsize $\pm$ 0.552} & 37.60 {\scriptsize $\pm$ 0.518} \\
      \hline
\end{tabular}
\end{small}
 \end{center}
\caption{\label{res} Average~$F_{1}$~scores and confidence intervals of methods applied on \mbox{EN$\rightarrow$FR} sub-corpora at chunk and sentence level -- 10 folds validation.}
\end{table*}

Table~\ref{meth_gran} shows the Pearson correlations of the results (of all methods on all sub-corpora) by language pair between the chunk and the sentence granularity (correlations calculated from Table~\ref{correl_lang}, between the EN$\rightarrow$FR column at chunk level with the EN$\rightarrow$FR column at sentence level, and so on).
We can see a strong Pearson correlation of the performances on the language pair between the chunk and the sentence granularity (an average of 0.9, with 0.907 for the EN$\rightarrow$FR pair, for instance).
This proves that all methods behave along a similar trend at chunk and at sentence level, regardless of the languages on which they are used.
However, we can see in Table~\ref{lang_gran} that if we collect correlation scores separately for each method (on all sub-corpora, on all language pairs) between chunk and sentence granularity performances (correlations also calculated from Table~\ref{correl_lang}, between the \emph{\mbox{CL-C3G}} line at chunk level with the \emph{\mbox{CL-C3G}} line at sentence level, and so on), we notice that some methods exhibit a different behavior at both chunk and sentence granularities: for instance, this is the case for \emph{\mbox{CL-ASA}} which seems to be really better at chunk level.
In conclusion, we can say that the methods presented here may behave slightly differently depending on the text unit considered (chunk or sentence) but they behave practically the same no matter the languages of the compared texts are (as long as enough lexical resources are available for dealing with these languages).

\subsection{Detailed Analysis for English-French}
\label{detailed}

The previous sub-section has shown a consistent behavior of methods across language pairs (strongly consistent) and granularities (less strongly consistent). 
For this reason, we now propose a detailed analysis for different sub-corpora, \textit{for the English-French language pair - at chunk and sentence level - only}. 
Providing these results for all language pairs and granularities would take too much space.
Moreover, we also run those state-of-the-art methods on the dataset of the Spanish-English cross-lingual Semantic Textual Similarity task of SemEval-2016 \cite{semeval2016} and SemEval-2017 \cite{semeval2017}, and propose a shallower but equally rigorous analysis.
However, all those results are also made available as supplementary material on our paper Web page.

Table~\ref{res} shows the performances of methods on the EN$\rightarrow$FR sub-corpora.
As mentioned earlier, \emph{\mbox{CL-C3G}} is in general the most effective method.
\emph{\mbox{CL-ESA}} seems to show better results on comparable corpora, like Wikipedia.
In contrast, \emph{\mbox{CL-ASA}} obtains better results on parallel corpora such as JRC or Europarl collections.
\emph{\mbox{CL-CTS}} and \emph{\mbox{T+MA}} are pretty efficient and versatile too.
It is also interesting to note that the results of the methods are well correlated between certain types of sub-corpora.
For instance, the Pearson correlation of the performances of all methods between the TALN sub-corpus and the APR sub-corpus, is 0.982 at the chunk level, and 0.937 at the sentence level.
This means that a method could be optimized on a particular corpus (for instance APR) and applied efficiently on another corpus (for instance TALN which is made of scientific conference papers).

\begin{figure*}[ht!]
    \centering
    \subfloat[Distribution histogram (fingerprint) of a random distribution.]{{\includegraphics[width=7.7cm]{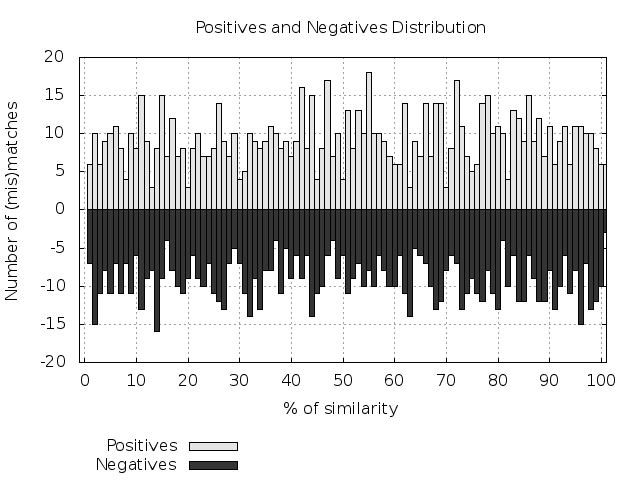} }}
    \quad
    \subfloat[Distribution histogram (fingerprint) of the \emph{\mbox{Length Model}} of \newcite{pouliquen2003a}.]{{\includegraphics[width=7.7cm]{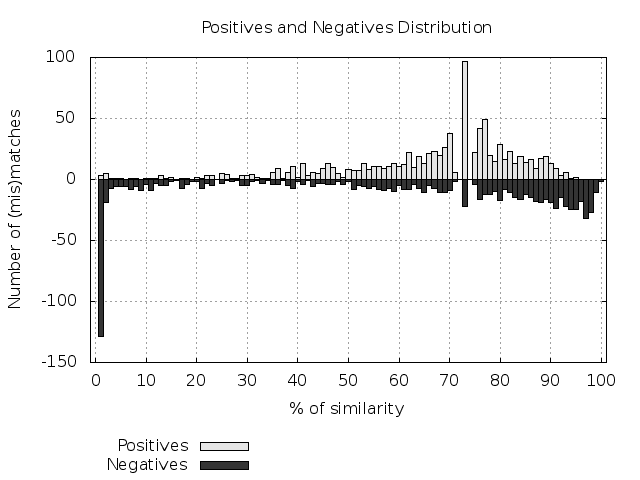} }}
    \\
    \subfloat[Distribution histogram (fingerprint) of \emph{\mbox{CL-C3G}}.]{{\includegraphics[width=7.7cm]{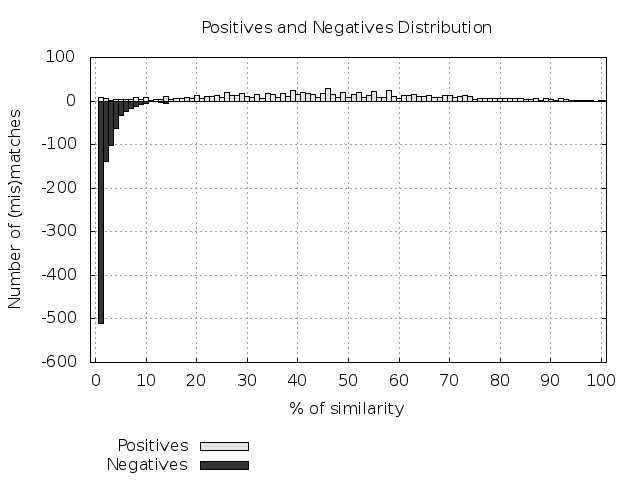} }}
    \quad
    \subfloat[Distribution histogram (fingerprint) of \emph{\mbox{CL-CTS}}.]{{\includegraphics[width=7.7cm]{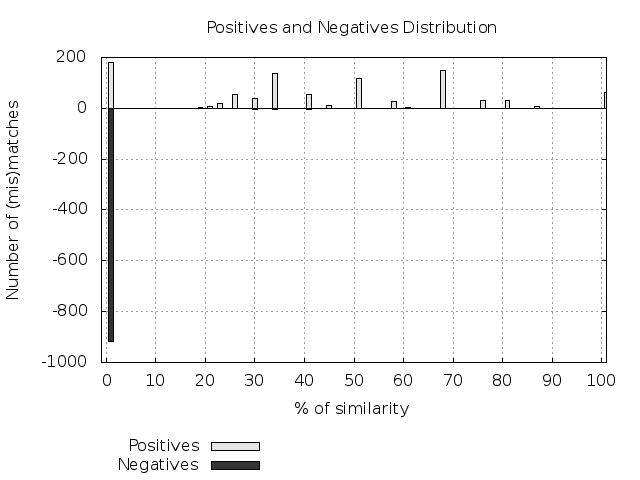} }}
    \\
    \subfloat[Distribution histogram (fingerprint) of \emph{\mbox{CL-ASA}}.]{{\includegraphics[width=7.7cm]{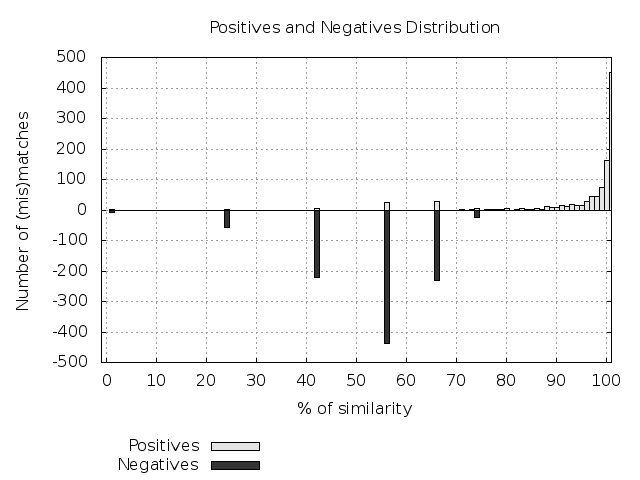} }}
    \quad
    \subfloat[Distribution histogram (fingerprint) of \emph{\mbox{T+MA}}.]{{\includegraphics[width=7.7cm]{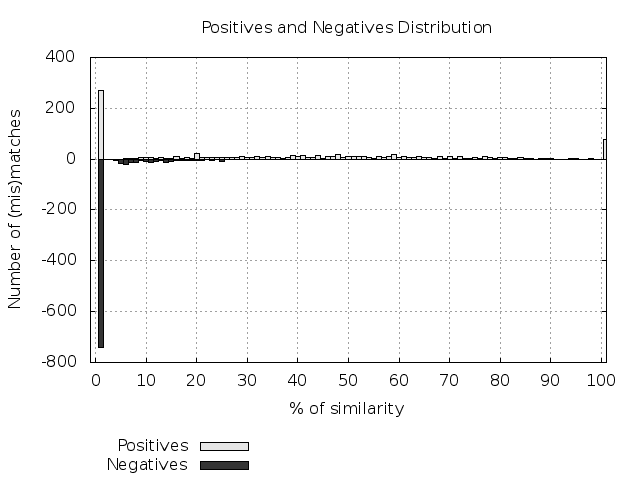} }}
    \caption{
Distribution histograms of some state-of-the-art methods for 1000 positives and 1000 negatives (mis)matches.
X-axis represents the similarity score (in percentage) computed by the method, and Y-axis represents the number of (mis)matches found for a given similarity score.
In white, in the upper part of the figures, the positives (units that needed to be matched), and in black, in the lower part, the negatives (units that should not be matched).
}
    \label{fingerprint}
    \vspace{-0.5cm}
\end{figure*}

Beyond their capacity to correctly predict a (mis)match, an interesting feature of the methods is their clustering capacity, \textit{i.e.} their ability to correctly separate the positives (cross-lingual semantic textual similar units) and the negatives (textual units with different meaning) in order to minimize the doubts on the classification.
To verify this phenomenon, we conducted another experience with a new protocol.
We built a data subset by concatenating some documents of the previously presented dataset \cite{dataset-lrec}. More precisely we used 200~pairs of each sub-corpora at sentence level only.
We compared 1000 English textual units to their corresponding unit in French, and to one other (not relevant) French unit.
So, each English textual unit must strictly leads to one match and one mismatch, \textit{i.e.} in the end, we have exactly 1000 matches and 1000 mismatches for a run.
We repeat this experiment 10 times for each method, leading to 10 folds for each method.

\begin{table}[t!]
 \begin{center}
\begin{small}
\begin{tabular}{|l|l|l|l|l|}
      \hline
      \bf Methods & \bf \textit{T} & \bf \textit{P} & \bf \textit{R} & \bf $F_{1}$ \\
      \hline
      Random baseline & 0.003 & 0.501 & 0.999 & 0.668 \\
      Length Model & 0.203 & 0.566 & 0.970 & 0.714 \\
      CL-C3G & 0.087 & 0.972 & 0.953 & 0.962 \\
      CL-CTS & 0.010 & 0.986 & 0.808 & 0.888 \\
      CL-ASA & 0.762 & 0.937 & 0.772 & 0.847 \\
      T+MA & 0.157 & 0.928 & 0.646 & 0.762 \\
      \hline
\end{tabular}
\end{small}
 \end{center}
\caption{\label{t_p_r} Precision (\textit{P}), Recall (\textit{R}) and $F_{1}$~score, reached at a certain threshold (\textit{T}), of some state-of-the-art methods for a data subset made with 1000 positives and 1000 negatives (mis)matches -- 10 folds validation.}
\end{table}

The results of this experiment are reported on Table~\ref{t_p_r}, that shows the average for the 10 folds of the Precision (\textit{P}), the Recall (\textit{R}) and the $F_{1}$~score of some state-of-the-art methods, reached at a certain threshold (\textit{T}).
The results are also reported in Figure~\ref{fingerprint}, in the form of distribution histograms of the evaluated methods for 1000 positives and 1000 negatives (mis)matches.
X-axis represents the similarity score (in percentage) computed by the method, and Y-axis represents the number of (mis)matches found for a given similarity score.
In white, in the upper part of the figures, the positives (units that needed to be matched), and in black, in the lower part, the negatives (units that should not be matched).

Distribution histograms on Figure~\ref{fingerprint} highlights the fact that each method has its own fingerprint:
even if two methods looks equivalent in term of performances (see Table~\ref{t_p_r}), their clustering capacity, and so the distribution of their (mis)matches can be different.
For instance, we can see that a random distribution is a very bad distribution \mbox{(Figure~\ref{fingerprint} (a))}.
We can also see that \emph{\mbox{CL-C3G}} has a narrow distribution of negatives and a broad distribution for positives \mbox{(Figure~\ref{fingerprint} (c))}, whereas the opposite is true for \emph{\mbox{CL-ASA}} \mbox{(Figure~\ref{fingerprint} (e))}.
Table~\ref{t_p_r} confirms this phenomenon by the fact that the decision threshold is very different for \emph{\mbox{CL-ASA}}~(0.762) compared to the other methods \mbox{(around~0.1)}.
This means that \emph{\mbox{CL-ASA}} discriminates more correctly the positives that the negatives, when it seems to be the opposite for the other methods.
For this reason, we can make the assumption that some methods are complementary, due to their different \textit{fingerprint}.
These behaviors suggest that fusion between these methods (notably decision tree based fusion) should lead to very promising results.

\section{Conclusion}
\label{conclusion}

We conducted a deep investigation of cross-language plagiarism detection methods on a challenging dataset. 
Our results have shown a common behavior of methods across different language pairs.
We revealed strong correlations across languages but also across text units considered. 
This means that when a method is more effective than another on a sufficiently large dataset, it is generally more effective in any other case.
This also means that if a method is efficient on a particular language pair, it will be similarly efficient on another language pair as long as enough lexical resources are available for these languages.

We also investigated the behavior of the methods through the different types of texts on a particular language pair: English-French.
We revealed strong correlations across types of texts.
This means that a method could be optimized on a particular corpus and applied efficiently on another corpus.

Finally, we have shown that methods behave differently in clustering match and mismatched units, even if they seem similar in performance.
This opens new possibilities for their combination or fusion.

More results supporting these facts are provided as supplementary material\footnote{\url{https://github.com/FerreroJeremy/Cross-Language-Dataset/tree/master/study}}.

\bibliography{bucc2017}
\bibliographystyle{acl_natbib}

\appendix

\end{document}